\documentclass[]{article}
\usepackage{proceed2e}
\usepackage{times}
\usepackage{booktabs}
\usepackage{stackrel}
\usepackage{subscript} 
\usepackage[geometry]{ifsym} 
\usepackage{color} 
\usepackage[utf8]{inputenc} 
\usepackage{amsmath,amssymb,amsthm} 
\usepackage{graphicx} 
\usepackage{subfigure} 
\usepackage{booktabs} 
\usepackage{multirow} 
\usepackage{multicol}
\usepackage[round]{natbib}
\usepackage{url}
\usepackage{caption}
\usepackage{paralist}
\usepackage[lined, ruled, boxed, linesnumbered, commentsnumbered]{algorithm2e}
\usepackage{titlesec} 
\titlespacing\section{0pt}{12pt plus 4pt minus 2pt}{0pt plus 2pt minus 2pt}
\titlespacing\subsection{0pt}{12pt plus 4pt minus 2pt}{0pt plus 2pt minus 2pt}
\titlespacing\subsubsection{0pt}{12pt plus 4pt minus 2pt}{0pt plus 2pt minus 2pt}

\newcommand{\argmin}[1]{\underset{#1}{\operatorname{arg}\!\operatorname{min}}\;}

\newcommand{\Do}[1]{do(#1)}
\newcommand{\man}[1]{man(#1)}
\newcommand{\elnino}[0]{El Ni\~{n}o}
\newcommand{\lanina}{La Ni\~{n}a}
\newcommand{\calX}{\mathcal{X}}
\newcommand{\calY}{\mathcal{Y}}

\newtheorem{theorem}{Theorem}

\newtheorem{definition}[theorem]{Definition}

\title{Unsupervised Discovery of \elnino{} \\ Using Causal Feature Learning on Microlevel Climate Data}

\author{ {\bf Krzysztof Chalupka} \\
Computation and Neural Systems\\
Caltech\\
\And
{\bf Tobias Bischoff} \\
Environmental Science \\
and Engineering\\
Caltech\\
\And
{\bf Pietro Perona}  \\
Electrical Engineering\\
Caltech\\
\And
{\bf Frederick Eberhardt}   \\
Humanities and Social Sciences\\
Caltech\\
}

\begin{document}

\maketitle

\begin{abstract}
We show that the climate phenomena of \elnino{} and \lanina{} arise naturally as states of macro-variables when our recent causal feature learning framework~\citep{Chalupka2015a,Chalupka2015b} is applied to micro-level measures of zonal wind (ZW) and sea surface temperatures (SST) taken over the equatorial band of the Pacific Ocean. The method identifies these unusual climate states on the basis of the relation between ZW and SST patterns without any input about past occurrences of \elnino{} or \lanina{}.
The simpler alternatives of (i) clustering the SST fields while disregarding their relationship with ZW patterns, or (ii) clustering the joint ZW-SST patterns, do not discover \elnino{}. We discuss the degree to which our method supports a causal interpretation and 
use a low-dimensional toy example to explain its success over other clustering approaches. Finally, we propose a new robust and scalable alternative to our original algorithm~\citep{Chalupka2015b}, which circumvents the need for high-dimensional density learning.
\end{abstract}

\section{INTRODUCTION}
The accurate characterization of macro-level climate phenomena is crucial to an understanding of climate dynamics, long term climate evolution and forecasting. Modern climate science models, despite their complexity, rely on an accurate and valid aggregation of micro-level measurements into macro-phenomena. While many aspects of the climate may indeed be subject fundamentally to chaotic dynamics, many large scale phenomena are deemed amenable to precise modeling. The \elnino{}--Southern Oscillation (ENSO) is arguably the most studied climate phenomenon at the inter-annual time scale, but much about its dynamics relating zonal winds (ZW) and sea surface temperatures (SST) remains poorly understood.

\begin{figure}[ht]
\centering
\includegraphics[width=0.485\textwidth]{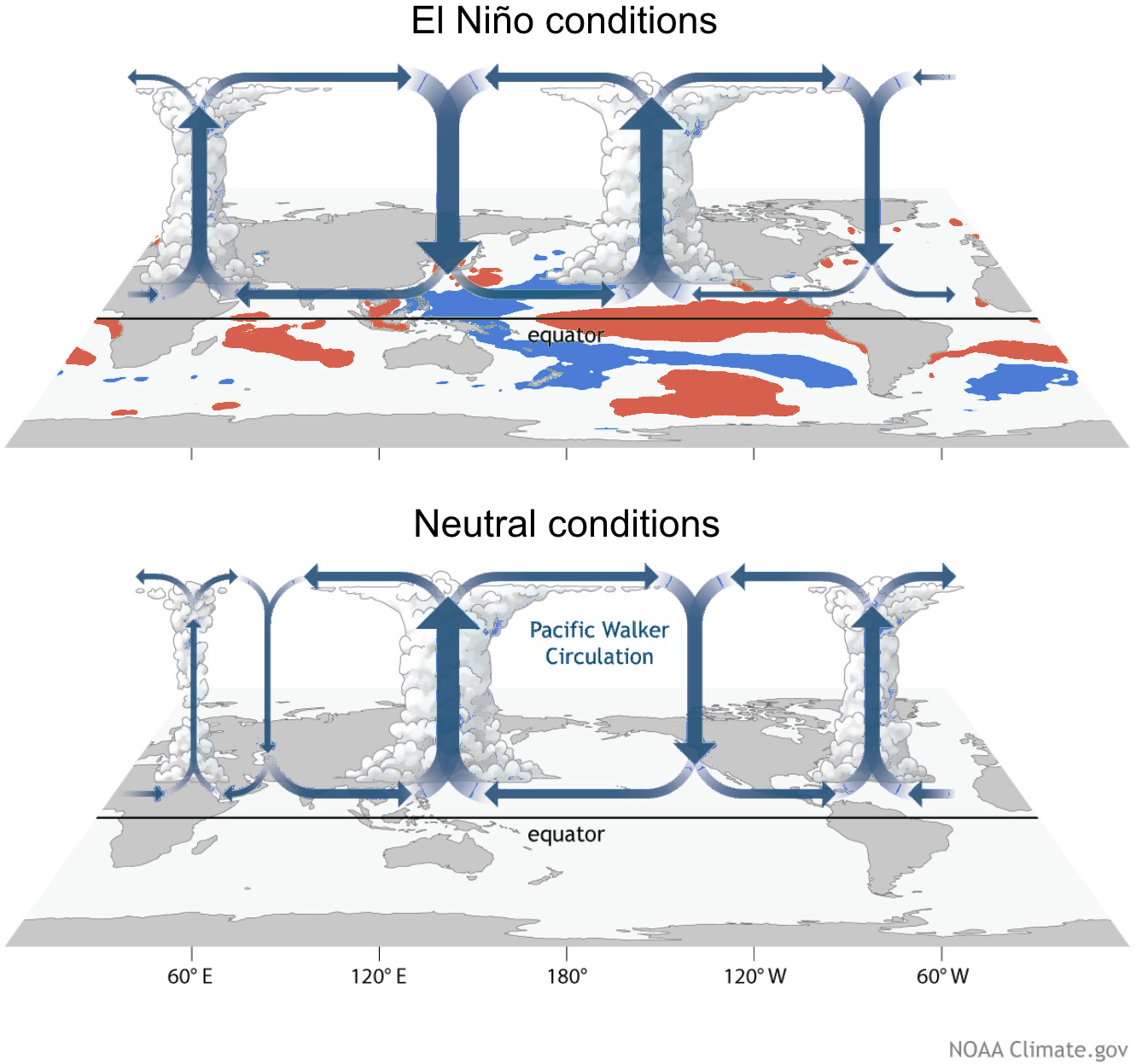}
\caption{\elnino{} vs. neutral conditions from \citet{diliberto2014}. Top: An illustration of the state of the atmosphere and surface during typical \elnino{} conditions. Here, the colors indicate SST deviations from the neutral state with red being a positive and blue being a negative deviation. Bottom: Similar to the top panel but now showing neutral conditions of the Walker circulation (neither \elnino{} nor \lanina{}).}
\label{fig:enso1}
\end{figure}

We apply our recent causal feature learning (CFL) framework~\citep{Chalupka2015b} to learn causal macro-variables from the equatorial Pacific climate data. Our goal is threefold:
\begin{compactitem}
\item apply CFL to real-world data, developing new practical algorithms as needed, 
\item test whether CFL can, without supervision, learn the ground truth that \elnino{} is an important macro-variable state in the ZW-SST system's dynamics,
\item explore the theoretical and practical difference between CFL and clustering methods. 
\end{compactitem}
From the climate-science point of view, our research shows that CFL can be successfully used for an unbiased automated extraction of climate macro-variables, which would otherwise require tedious hand-crafting by domain experts. Moreover, the framework can directly suggest (computationally) expensive climate experiments (for example, through climate simulations) that could differentiate between true causes and mere correlations efficiently. Closer inspection of the output of CFL can also yield insights about new climate macro-phenomena (or important variants of existing ones) that inspire new physical models of the climate. Python code that reproduces our results and figures is available online at \url{http://vision.caltech.edu/~kchalupk/code.html}.

\subsection{EL NI\~{N}O--SOUTHERN OSCILLATION}
\label{sec:elnino}
\elnino{} is a weather pattern that is principally characterized by the state of eastern Pacific near-surface winds (ZW, zonal wind), sea surface temperature (SST) patterns, and the associated state of the atmospheric Walker circulation~\citep[see for example,][]{holton1989, trenberth1997}. The Walker circulation (see Fig.~\ref{fig:enso1}) is characterized by warm air rising over Indonesia and Papua New Guinea and cooler subsiding air over the eastern Pacific cold tongue region just west of equatorial South America \citep{lau2003}. Near the surface, easterly winds (winds blowing from the east) drive water from east to west resulting in oceanic upwelling near the coast of equatorial South America (and downwelling east of Indonesia), that brings with it cold and nutrient rich waters from the deep oceans. During the ENSO warm phase, commonly referred to as \elnino{} (because it often occurs around and after Christmas), the Walker circulation weakens, ultimately resulting in weaker upwelling in the Eastern Pacific and thus in positive SST anomalies. Fig.~\ref{fig:enso1} illustrates these phenomena. 

\begin{figure}
\centering
\includegraphics[width=0.485\textwidth]{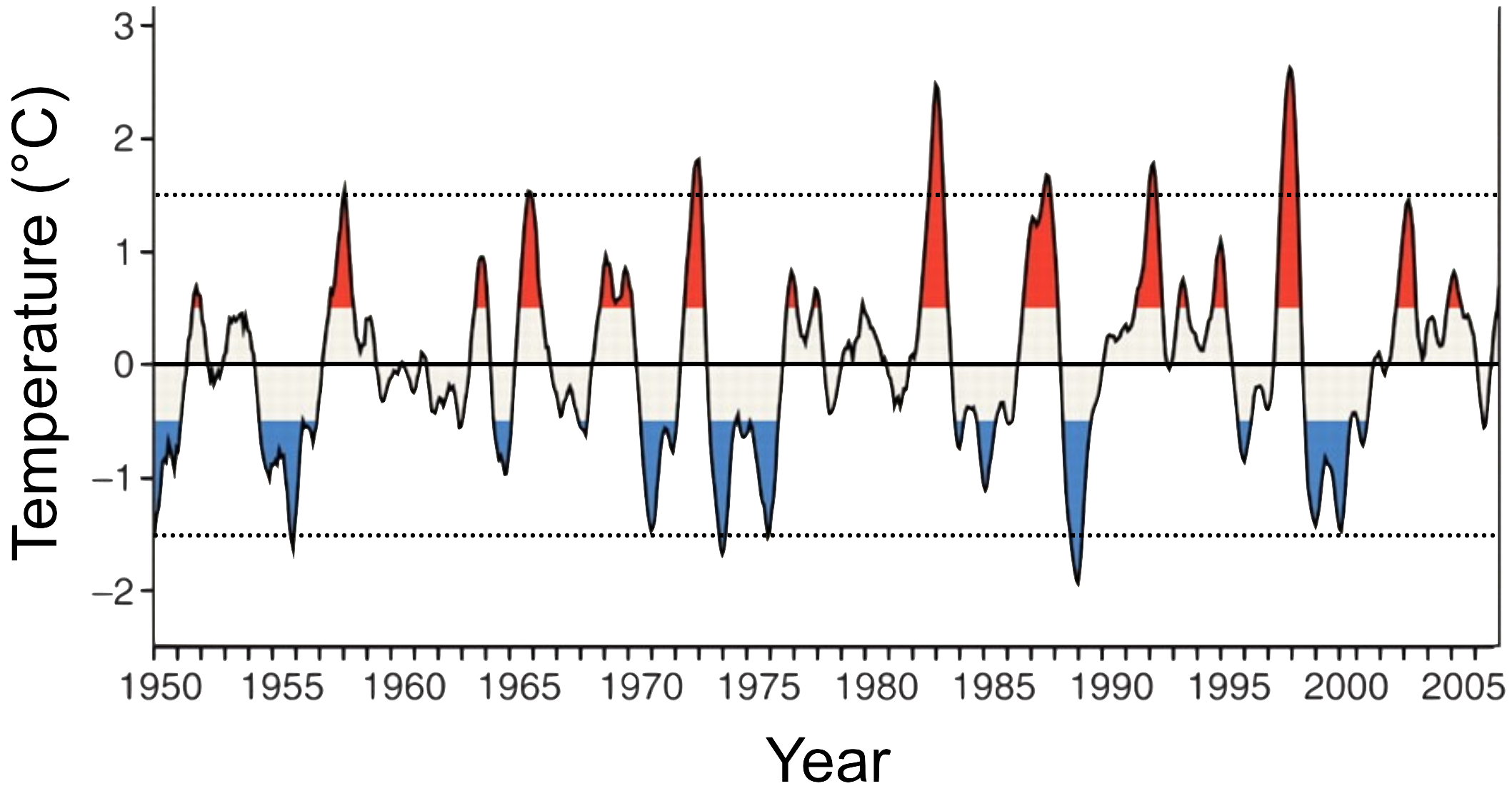}
\caption{Ni\~{n}o 3.4 SST anomalies for the time period 1950--2005. The figure was adapted from \cite{mcphaden2006}. Red shadings indicate \elnino{} years and blue shadings indicate \lanina{} years. The two dashed lines indicate the threshold for strong \elnino{} or \lanina{} events.}
\label{fig:enso2}
\end{figure}

ENSO-related weather in the tropics includes droughts, flooding, and may have direct impact on fisheries through reduced nutrient upwelling \citep[e.g.,][]{glantz2001}. Atmospheric waves (ripples in wind, SST and rainfall patterns) generated by the change in circulation and SST anomalies in the tropics, make their way across the planet with dramatic impact~\citep[e.g,][]{ropelewski1987, changnon1999}. \cite{cashin2015} show that the economic impact of \elnino{} varies across regions. Economic activity may decline briefly in Australia, Chile, Indonesia, India, Japan, New Zealand, and South Africa after an \elnino{} event. Enhanced growth may be registered in other countries, such as the United States.

The ENSO cold phase, usually referred to as \lanina{}, is the opposing phase of \elnino{} with enhanced upwelling and colder SSTs in the eastern Pacific. Currently, predicting the strength of \elnino{} and \lanina{} events remains a difficult challenge for climate scientists as the period may vary between 3 and 7 years (see Fig.~\ref{fig:enso2}); as a consequence accurate forecasts are only possible less than a year in advance \citep[e.g.,][]{landsea2000}.

\begin{figure}
\centering
\includegraphics{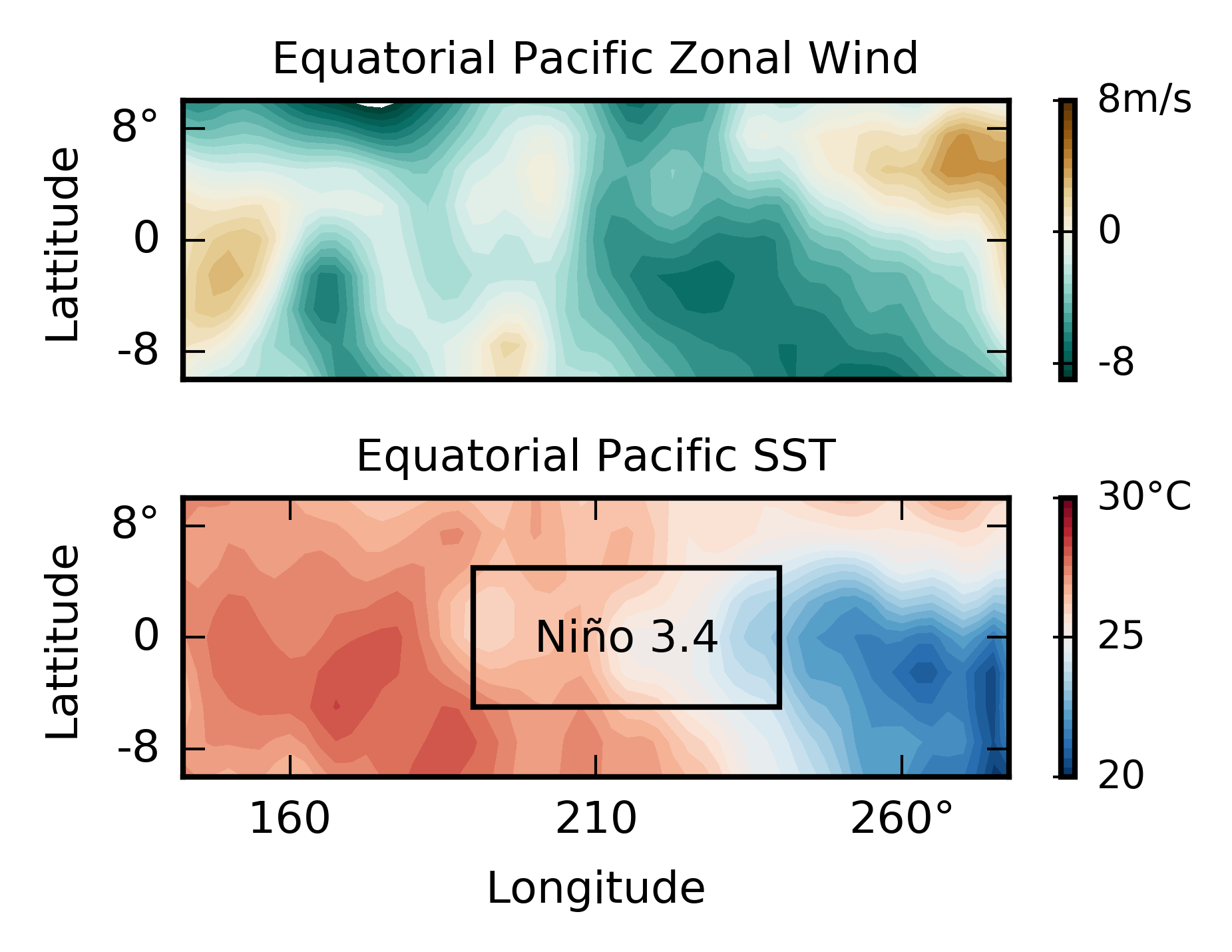}
\caption{A micro-variable climate dataset. Top: A week's average ZW field. Bottom: A week's average SST field over the same region. In addition, the Ni\~{n}o 3.4 region is marked. Our dataset comprises 36 years' worth of overlapping weekly averages over the presented region.}
\label{fig:data}
\end{figure}

The National Oceanic and Atmospheric Administration (NOAA) defines \elnino{} as a positive three-month running mean SST anomaly of more than $0.5^\circ$C from normal (for the 1971--2000 base period) in the Ni\~{n}o 3.4 region ($120^\circ$W--$170^\circ$W, $5^\circ$N--$5^\circ$S, see also Fig.~\ref{fig:data}). Similarly, \lanina{} conditions are defined as negative anomalies of more than $-0.5^\circ$~C. Conditions in between $-0.5^\circ$C and $0.5^\circ$C are called neutral. This is illustrated using red and blue shadings in Fig.~\ref{fig:enso2}. Strong \elnino{}/\lanina{} events are defined as SST-anomalies greater than $1.5^\circ$C. However, the definitions for \elnino{} and \lanina{} have evolved over time. For example, other regions than the Ni\~{n}o 3.4 region or other averaging conventions have been used in the specification of the SST anomalies.

\subsection{CAUSAL FEATURES AND MACRO-VARIABLES}
Climate experts view zonal winds as drivers of SST patterns. We take the view that 
if \elnino{} and \lanina{} are indeed genuine macro-level climate phenomena in their own right (and not just arbitrary quantities defined by convention) then they must consist of macro-level features of the relation between the high-dimensional micro-level ZT and SST patterns that can be detected by an unsupervised method. That is, it must be possible to identify \elnino{} and \lanina{} from a mass of air pressure and sea temperature readings, using a method that has no independent information about when such periods occurred.

In \citet{Chalupka2015b} we developed a theoretically precise account of causal relations of macro-variables that supervene on micro-variables, and proposed an unsupervised method for their discovery, which we called Causal Feature Learning (CFL). We adopt the framework (summarized below) with a few interpretational adjustments for our climate setting. The method (originally inspired by the neuroscience setting, only tested on synthetic data) was designed to establish claims such as \emph{``The presence of faces (in an image) causes specific neural processes in the brain."}, where a neural process identifies a class of spike trains across a large number of neurons recorded by electrodes. An ability to characterize such neural processes would provide the basis to explain, for example, what constitutes face recognition in the brain. There we considered as input visual stimuli (in the form of still images) and as output electrode recordings of the neural response of 1000 neurons (in the form of spike trains).   

Formally, let an input (micro-)variable $X$ take values in a  high-dimensional domain $\calX$ (in \citet{Chalupka2015b}, the pixel space of an image, in our case here ZW maps) and the output \mbox{(micro-)}variable $Y$ take values in the high-dimensional domain $\calY$ (the space of neural spike trains then, the SST patterns here). The basic idea underlying our set-up is that the causal macro-variable relation is defined in terms of the \emph{coarsest} aggregation of the micro-level spaces that preserves the probabilistic relations under intervention (hence, causal) between the micro-level spaces. Conceptually, macro-level causal variables group together micro-level states that make no causal difference. In \citet{Chalupka2015b} we started by defining a micro-level manipulation (similar to Pearl's $\Do$-operator \citep{Pearl2000}):

\begin{definition}[Micro-level Manipulation]
A \emph{micro-level manipulation} is the operation $\man{X=x}$ that changes the value of the micro-variable $X$ to $x\in\calX$, while not (directly) affecting any other variables. We write $\man{x}$ if the manipulated variable $X$ is clear from context.
\end{definition}\label{def:fcp}

The micro-level manipulation is then used to define what we refer to as the \emph{fundamental causal partition}:

\begin{definition}[Fundamental Causal Partition, Causal Class]
\label{def:macrovars}
Given the pair $(\calX, \calY)$, the \emph{fundamental causal partition of $\calX$}, denoted by $\Pi_c(\calX)$ is the partition induced by the equivalence relation $\overset{X}\sim$ such that 
\begin{align*}
x_1 \overset{X}{\sim} x_2 & \quad \Leftrightarrow \quad \forall_{y}\; P(y\mid \man{x_1}) = P(y\mid \man{x_2}).
\end{align*}
Similarly, the \emph{fundamental causal partition of $\calY$}, denoted by $\Pi_c(\calY)$, is the partition induced by the equivalence relation $\overset{Y}\sim$ such that 
\begin{align*}
y_1 \overset{Y}{\sim} y_2 \quad \Leftrightarrow \quad \forall_{x}\;P(y_1\mid \man{x}) = P(y_2 \mid \man{x}). 
\end{align*}
A cell of a causal partition is a \emph{causal class} of $X$ or $Y$.
\end{definition}
The fundamental causal partitions then naturally give rise to the macro-level cause variable $C$ and effect variable $E$ that stand in a bijective relation to the cells of $\Pi_c(\calX)$ and $\Pi_c(\calY)$, respectively. Thus, the macro-variable cause $C$ ignores all the micro-level changes in $X$ that do not have an effect on the probabilities over $Y$, and the macro-level effect $E$ ignores all the micro-level detail in $Y$, which occur with the same probability given a manipulation to any $X=x$. 

\begin{figure}
\centering
\includegraphics{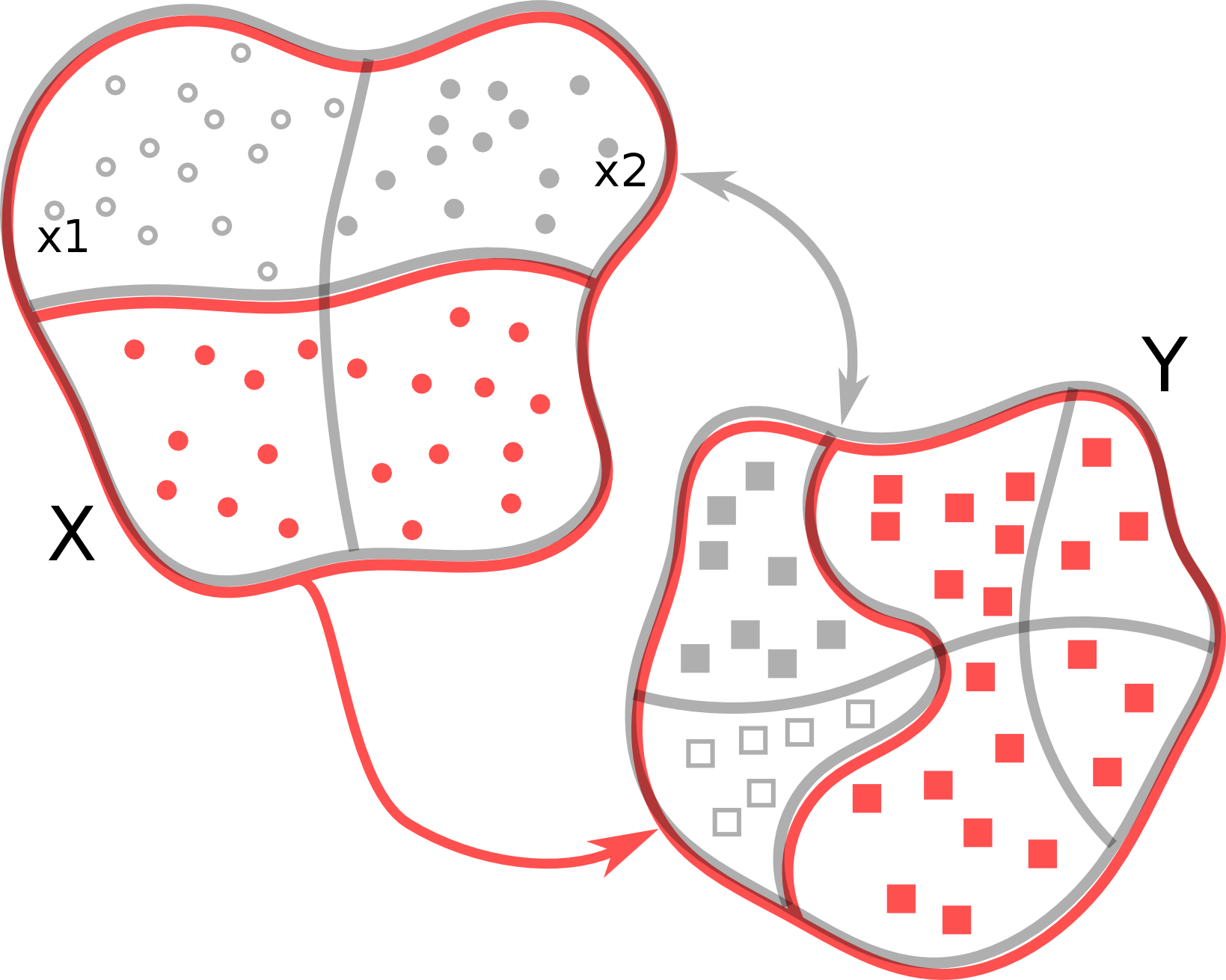}
\caption{The Causal Coarsening Theorem, adapted from \citet{Chalupka2015b}. In this plot, the \emph{observational} input macro-variable (top, gray) has four states, and has a well-defined joint with the observational output macro-variable (with six states). In each case, the \emph{causal} macro-variable states are a coarsening of the observational states. For example, the input causal macro-variable merges the two top observational states. E.g.\ $P(Y\mid x_1) \neq P(Y\mid x_2)$, but 
$P(Y\mid \man{x_1})=P(Y\mid\man{x_2})$.}
\label{fig:cct}
\end{figure}

With these definitions there is no reason \emph{a priori} to think that macro-variables are common phenomena. In fact quite the opposite: The conditions that the probability distributions over $X$ and $Y$ must satisfy to give rise to non-trivial macro-variables $C$ and $E$ can easily be described as a measure-zero event when taken in their strict form. Consequently, our view is that to the extent that macro-variables are discussed in a scientific domain, there must be a pre-supposition that such strong conditions are satisfied at least approximately.

In the present context, our climate data consisting of ZW and SST measurements (we give a detailed description of the data in Section~\ref{sec:data} below) is entirely observational. That is, the data is naturally sampled from $P(\text{SST},\;\text{ZW})$ and not created by a (hypothetical) experimentalist from $P(\text{SST} \mid \man{\text{ZW}=z})$ for different values of $z$. Nevertheless, we can identify the \emph{observational} macro-variables that characterize the probabilistic relation between ZW and SST by replacing the probabilities in Definition~\ref{def:fcp} with observational probabilities $P(y\mid x)$: 
\begin{definition}[Fundamental Observational Partition, Observational Class]
\label{def:macrovars2}
Given the pair $(\calX, \calY)$, the \emph{fundamental observational partition of $\calX$}, denoted by $\Pi_o(\calX)$ is the partition induced by the equivalence relation $\overset{X}\sim$ such that 
\begin{align*}
x_1 \overset{X}{\sim} x_2 & \quad \Leftrightarrow \quad \forall_{y}\; P(y\mid x_1) = P(y\mid x_2).
\end{align*}
Similarly, the \emph{fundamental observational partition of $\calY$}, denoted by $\Pi_o(\calY)$, is the partition induced by the equivalence relation $\overset{Y}\sim$ such that 
\begin{align*}
y_1 \overset{Y}{\sim} y_2 \quad \Leftrightarrow \quad \forall_{x}\;P(y_1\mid x) = P(y_2 \mid x). 
\end{align*}
A cell of an observational partition is an \emph{observational class} of $X$ or $Y$.
\end{definition}

In~\citet{Chalupka2015b} we showed that the  fundamental \emph{causal} partition is almost always a \emph{coarsening} of the corresponding fundamental \emph{observational} partition, as illustrated in Fig.~\ref{fig:cct}. We thus have some reason to expect that any macro-variables we do identify from our observational climate data will capture all the distinctions that are causal, but may in addition make some distinctions that do not support a causal inference. We return to this point in Section~\ref{sec:discussion}, where we discuss in more detail what causal insights can be drawn from this work. Our results should be seen as a step towards a characterization of macro-level causal variables for climate science, but we fully acknowledge that a complete causal characterization of the equatorial Pacific climate dynamics is beyond the scope of this paper.

\subsection{DATASET}
\label{sec:data}
The data used for this study is based on the daily-averaged version of the NCEP-DOE Reanalysis 2 product for the time period 1979--2014 inclusive \citep{kanamitsu2002}, a data product provided by the US National Centers for Environmental Protection (NCEP) and the Department of Energy (DOE). Reanalysis data sets are generated by fitting a complex climate model to all available data for a given period of time, thus generating estimates for times and locations that were not originally observed. In addition, we used the Geophysical Observational Analysis Tool (http://www.goat-geo.org) to interpolate the SST and zonal wind fields onto a $2.5^\circ \times 2.5^\circ$ spatial grid for easier analysis. We chose to focus on the \mbox{(140$^\circ$,\;\;280$^\circ$)E}$\times$\mbox{(-10$^\circ$,\;\;+10$^\circ$)N} equatorial band of the Pacific Ocean. From the raw dataset, we extracted the zonal (west-to-east) wind component and SST data in this region (specifically, we extracted the fields at the $1000$~hPa level near the surface). Finally, we smoothed the data by computing a running weekly average in each domain. The resulting dataset contains 13140 zonal wind and 13140 corresponding SST maps, each a 9$\times$55 matrix. Fig.~\ref{fig:data} shows sample data points.

\section{PACIFIC MACRO-VARIABLES}
To apply CFL in practice, we adapted our unsupervised causal feature learning algorithm~\citep{Chalupka2015b} to more realistic scenarios. The new solution (Sec.~\ref{sec:alg}) is more robust and applicable to high-dimensional real-world data. We start with a description of the results. 

\begin{figure}
\centering
\includegraphics{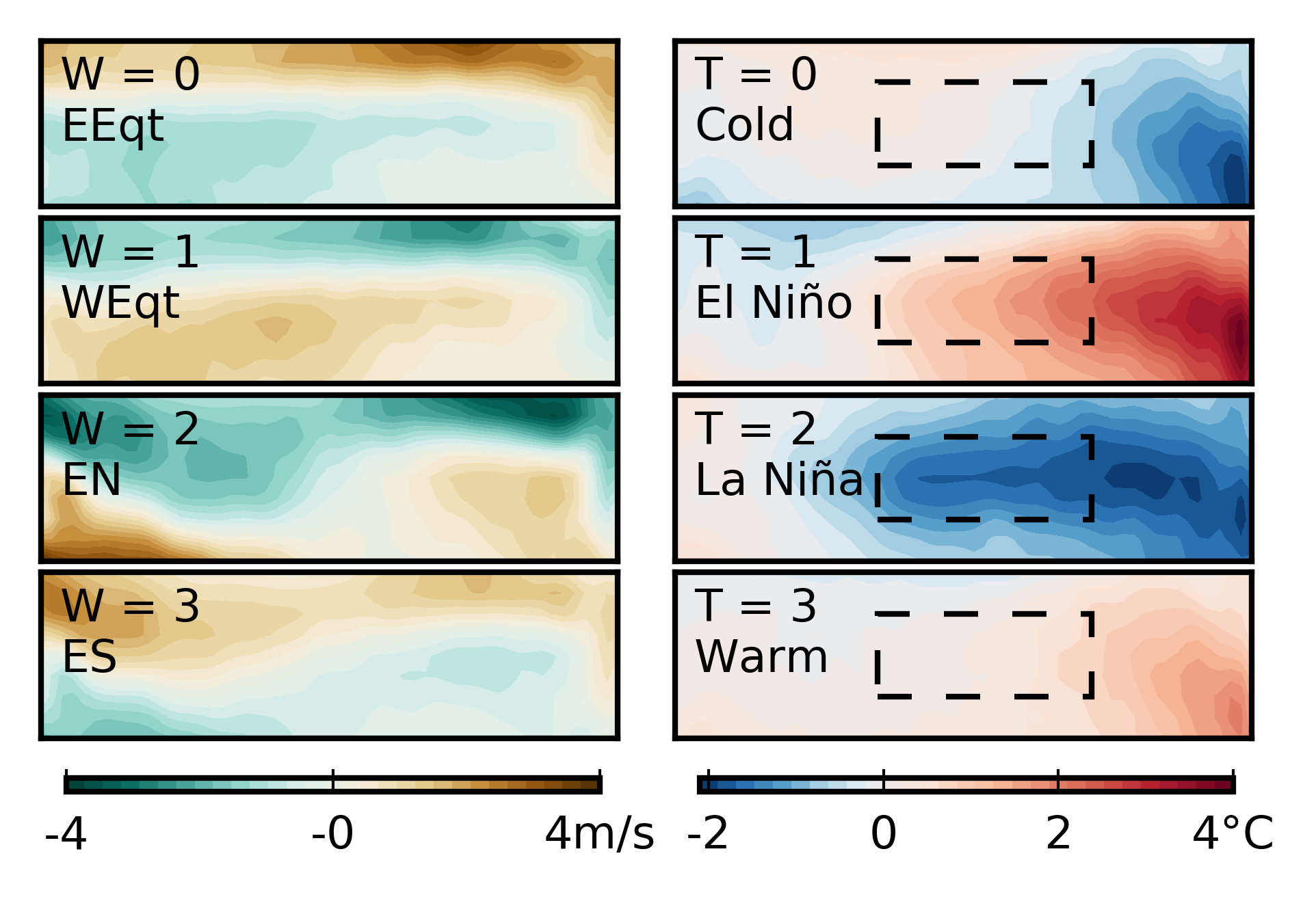}
\caption{Macro-variables discovered by Alg.~\ref{alg:cfl}. For each state, the average difference from the dataset mean is shown.  Left: Four states of W, the zonal wind macro-variable. We named the states ``Easterly Equatorial" (EEqt),``Westerly Equatorial" (WEqt), ``Easterly North of Equator" (EN) and ``Easterly South of Equator" (ES). Right: Four states of T, the SST macro-variable. We named the states ``Cold [American Coastal Waters]", ``\elnino{}", ``\lanina{}" and ``Warm [American Coastal Waters]". The main text provides additional justification for calling T=1 and T=2 ``\elnino{}" and "\lanina{}", respectively.}
\label{fig:cfl1}
\end{figure}

Throughout the article, we will refer to zonal wind \emph{macro-variables} as W, and to temperature \emph{macro-variables} as T. We first chose to search for four-state macro-variables (though we experiment with varying this number in Sec.~\ref{sec:varying_clusters}) and considered a zero-time delay\footnote{A zero time delay implies that CFL will attempt to relate the weekly moving ZW average to the weekly moving SST average. The question of different time delays turns out to be a very subtle issue in the study of \elnino{} as \elnino{} is not a periodic event, nor does it have a fixed duration (see Fig.~\ref{fig:enso2}). A careful discussion of other delays is not feasible in a short article and the zero-time delay was deemed a reasonable starting point by domain experts we consulted.} between W and T. In the CFL framework, each macro-variable state corresponds to a cell of a partition of the respective micro-variable input space. Fig.~\ref{fig:cfl1} visualizes the W and T we learned by plotting the difference between each macro-variable cell's mean and the ZW (SST) mean across the whole dataset. The visualized states are easy to describe: For example, when W=WEqt there is a larger-than-average westerly wind component in the west-equatorial region, a feature often associated with the causes of \elnino{} (see Fig.~\ref{fig:enso1}). Indeed, Table~\ref{tab:probtab} shows that the \elnino{} cell of T only arises in connection with W=WEqt. In addition, WEqt is often positively correlated with the T=Warm. Throughout the rest of the article, we will mostly focus on the T macro-variable. Our first goal is to quantitatively justify calling T=1 ``\elnino{}'' and calling T=2 ``\lanina{}''. Qualitatively, the warm and cold water tongues that reach westward across the Pacific and that are often used to describe the two phenomena, are evident in the image.

Following the standard definition of \elnino{} (see Section~\ref{sec:elnino}), we use the SST anomaly in the Ni\~{n}o 3.4 region to detect its presence~\citep{trenberth1997}. The anomaly is computed with respect to the climatological mean, that is the mean temperature \emph{during the same week of the year} over all the weeks in our dataset. We will call a weekly average anomaly exceeding +.5$^\circ$C a mild episode, and an anomaly exceeding +1.5$^\circ$C a strong episode. The definition of \lanina{} is analogous, with negative thresholds. Fig.~\ref{fig:cfl2} shows that in the T=1 and T=2 cells, over 75\% of all the points exceed the threshold for a mild (positive and negative, respectively) anomaly, and over $50\%$ of the points exceed the strong threshold. The situation is different in the Warm and Cold cells, where almost no points exceed the strong threshold while the number of points falling in these non-anomalous cells is about 30\% of the total. Since this macro-variable contains a state capturing a high proportion of \elnino{}-like patterns, we will say that this state has a ``high precision'' of detecting \elnino{}, while similarly, state T=2 has a high \lanina{} precision. Formally, we define the precision of a macro-variable state as follows:
\begin{figure}
\centering
\includegraphics{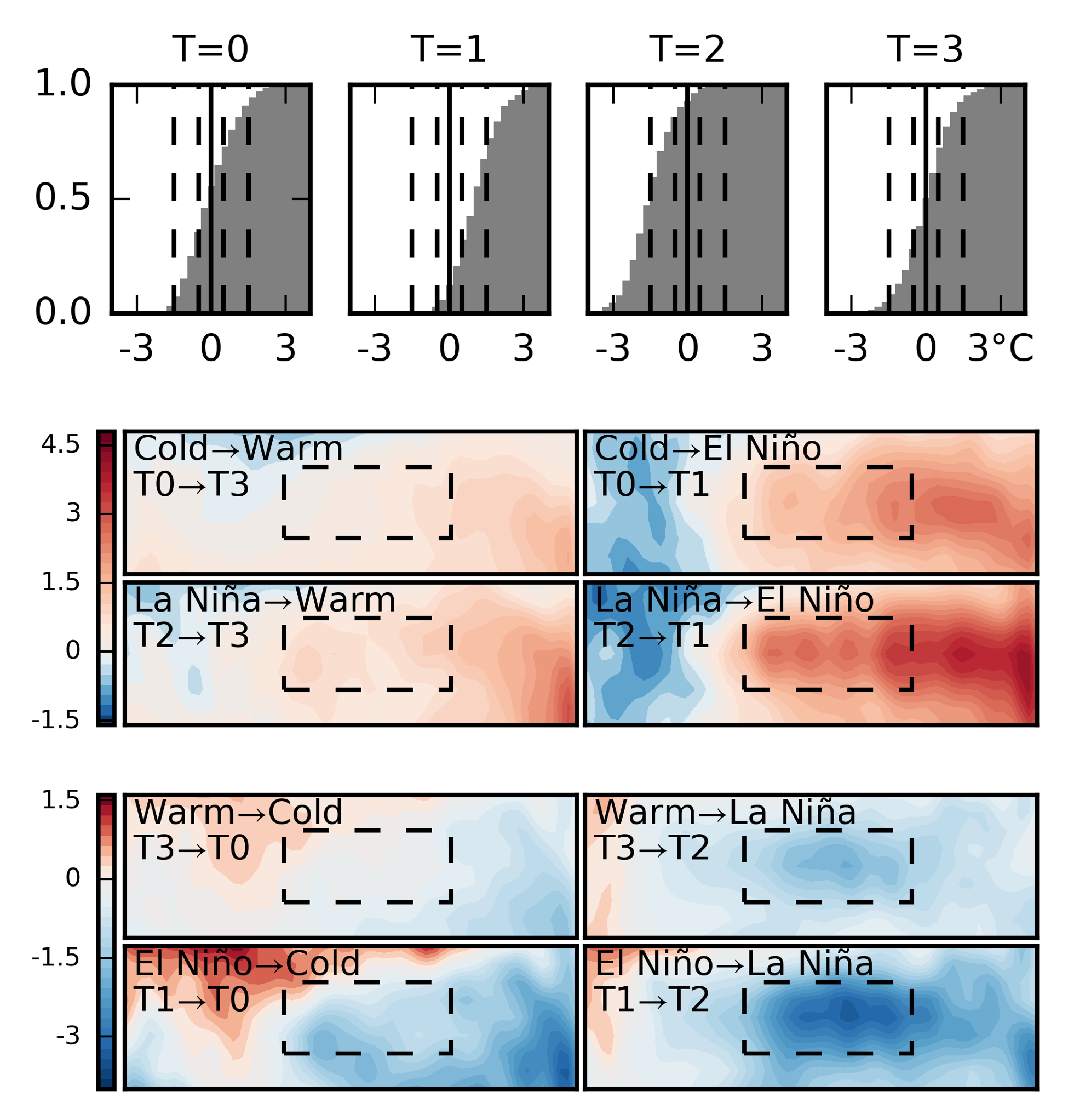}
\caption{T=1 and T=2 are \elnino{} and \lanina{}. Top: Each plot shows the cumulative histogram of the Ni\~{n}o 3.4 anomalies, computed over all the weekly SST averages that belong to the given state of T. The dashed lines show the +/-0.5 and +/-1.5 ``mild'' and ``strong'' anomaly thresholds. Bottom: The minimal manipulations needed to transition from a given T-state into another (the exact procedure to obtain the plots is described in the text).}
\label{fig:cfl2}
\end{figure}

\begin{definition}[precision]
Let $T=\{T_1, \cdots, T_K\}$ be a partition of the set of all the SST maps used in our experiments. Let $n34:\: SST\rightarrow \mathbb{R}$ be the function that computes the Ni\~{n}o 3.4 anomaly for a given map. Then, let
\[
c_\theta(T_k) = \left\{\begin{array}{l}
                \frac{1}{|T_k|}|\{t\in T_k\;s.t.\; n34(t)> \theta\}|\quad\text{if }\theta>0\\
                \phantom{W}\\
                \frac{1}{|T_k|}|\{t\in T_k\;s.t.\; n34(t)< \theta\}|\quad\text{if }\theta<0
                \end{array}\right.
\]
be the function that computes for, a given cell $T_k$ of the partition, the fraction of its members whose anomaly is greater than (if $\theta > 0$) or lesser than (if $\theta < 0$) a given threshold $\theta$. Finally, call the four numbers $\max_kc_{.5}(T_k)$, $\max_kc_{1.5}(T_k)$, $\max_kc_{(-.5)}(T_k)$, $\max_k c_{(-1.5)}(T_k)$ the mild/strong-\elnino{} and mild/strong-\lanina{} precision of the macro-variable $T$.
\label{def:enso_coeff}
\end{definition}

Together, the precisions indicate how well the partition T separates the mild and strong \elnino{} and \lanina{} anomalies from other structures in the data. In Fig.~\ref{fig:cfl2}, for example, $c_{.5}(T) \approx .75$ and $c_{1.5}(T) \approx .25$ (both because of T=1), $c_{(-.5)}(T) \approx .85$ and $c_{(-1.5)}(T) \approx .5$ (both because of T=2). Thus, T has high mild-\elnino{} precision, and high mild-\lanina{} precision.

As further evidence that Alg.~\ref{alg:cfl} recovered \elnino{} and \lanina{}, we show minimal state-to-state manipulations in Fig.~\ref{fig:cfl2}. Take the \lanina{}$\rightarrow$\elnino{} plot as an example. To compute it, we took all the SST maps for which T=\lanina{}, and for each found \emph{the closest} (in the Euclidean space) map for which T=\elnino{}. We then averaged these differences. One of the insights the figure offers is that low SSTs in the Ni\~{n}o 3.4 region really are the distinguishing feature of T=\lanina{}. Similarly, an important difference between the T=Warm and T=\elnino{} is the characteristic tongue of warm water extending into the Ni\~{n}o 3.4 region. Adding this tongue is necessary to switch from T=Cold to T=\elnino{}, but not to switch from T=Cold or T=\lanina{} to T=Warm.

The CFL framework allows us to interpret W and T as  standard probabilistic random variables with distribution we can estimate. Table~\ref{tab:probtab} offers a probabilistic description of the system we learned. ``When the equatorial zonal wind is unusually westerly, there is a 75\% chance that the eastern Pacific is warm, and a 25\% chance that \elnino{} arises.'' and ``When the North-equatorial zonal wind is predominantly westerly, but the South-equatorial easterly, then the Eastern Pacific is most likely to be cold.''---are example insights about the equatorial Pacific wind-SST system offered by CFL. We emphasize that both the macro-variables and the probabilities are learned from the data in an entirely unsupervised manner, without any a priori input about what constitutes ENSO events (except the fact that we restrict the SST and ZW fields to the equatorial Pacific region).

\section{CFL: A ROBUST ALGORITHM}
\label{sec:alg}
The practical bottleneck of the original CFL algorithm~\citep{Chalupka2015b} is the need for joint density estimation of $p(X,\;Y)$. Density estimation is notoriously hard, especially in high dimensions. We modified the original algorithm to avoid explicit density estimation. An additional advantage of our approach (Alg.~\ref{alg:cfl}) is that it is very robust with respect to input space dimensionality: Input data is only used explicitly in regression, which can be implemented using any algorithm that easily handles high-dimensional inputs (we used neural nets). 
\begin{figure}
\centering
\includegraphics{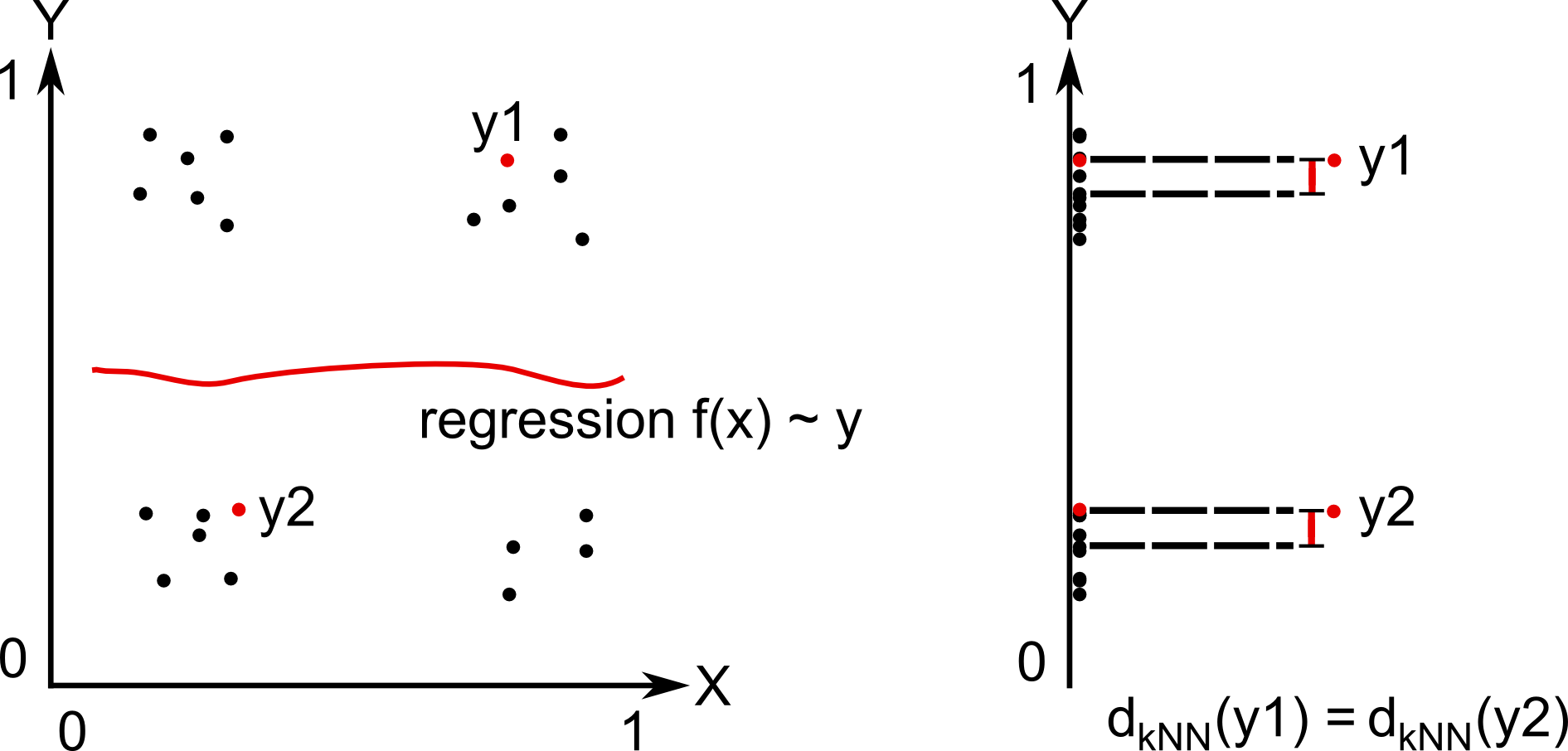}
\caption{Alg.~\ref{alg:cfl} vs.\ clustering. In this toy example, the data is sampled from the distribution $P(X) = U(\{1/5; 2/5)\}\cup\{3/5; 4/5\})$, $P(Y\mid X) = P(Y)$ $ = U(\{1/5; 2/5)\}\cup\{3/5; 4/5\})$. The clusters in the $\calX$, $\calY$, and joint $\calX,\calY$ space are evident. However, since $X$ and $Y$ are independent, we expect Alg.~\ref{alg:cfl} to find only one macrolevel class of $X$. Indeed, (properly regularized) regression gives $f(x)=const\;\forall_x$, so $W(x)=0\;\forall_x$. Incidentally, since the density of $Y$ is similar in the neighborhood of each sample $y$ (see data Y-projection on the right), $T(y)=0\;\forall_y$. }
\label{fig:c_vs_cfl}
\end{figure}

Let $\calX, \calY$ denote the micro-variable input and output space, respectively. Our algorithm is based on the insight that CFL only needs to detect the two equivalences
\begin{align}
p(Y \mid x_1) &= p(Y \mid x_2) \text{ for any } x_1, x_2\in \mathcal{X}\text{ and }\label{eq:x_eq} \\
p(y_1 \mid x) &= p(y_2 \mid x)\text{ for any } y_1, y_2 \in \mathcal{Y}, x\in \mathcal{X},\label{eq:y_eq}
\end{align}
instead of actually computing the conditionals $p(Y \mid X)$.

If Eq.~\eqref{eq:x_eq} holds, we also have $\mathbb{E}[Y \mid x_1] = \mathbb{E}[Y\mid x_2]$. Computing conditional expectations is much easier than learning the full conditional: $f(X) = \mathbb{E}[Y\mid X]$ minimizes $\mathbb{E}[(Y - f(X))^2]$, so learning the conditional expectation amounts to regressing $Y$ on $X$ under the mean-squared error measure. Unfortunately, equal conditional expectations do not imply equal conditional distributions. However, arguably the practical risk of encountering differing conditionals with identical means is lower than the risk of failing at high-dimensional density learning. For this reason, we use $\mathbb{E}[Y\mid x_1] = \mathbb{E}[Y\mid x_2]$ as a heuristic indicator of the equivalence of the conditionals in Eq.~\eqref{eq:x_eq} (see Line~\ref{alg:cfl1} in Alg.~\ref{alg:cfl}). For a more robust heuristic one could use more than just equal expectations to decide distribution equality. A promising direction would be to use a Mixture Density Network~\citep{Bishop1994} to approximate $P(Y \mid x)$ with a mixture of Gaussians for each $x$, and then cluster the mixtures. 

\IncMargin{1.5em}
\begin{algorithm}[t!]
\caption{Unsupervised Causal Feature Learning}
\label{alg:cfl}
\SetKwFunction{regress}{Regress}
\SetKwFunction{cluster}{Cluster}
\SetKwFunction{knn}{kNN}
\SetKwData{Data}{data}\SetKwData{NN}{NN}
\SetKwInOut{Input}{input}
\SetKwInOut{Output}{output}

\Input{$\mathcal{D} = {\{(x_1, y_1)},\cdots,(x_N, y_N)\}$\\
       $\cluster$ -- a clustering algorithm}
\Output{$W(x), T(y)$ -- the causal class of each $x, y$.}
\BlankLine
Regress $f \leftarrow \argmin{}\!_f\,\Sigma_i(f(x_i)-y_i)^2$\;
Let $W(x_i) \leftarrow \cluster(f(x_1), \cdots, f(x_N))[x_i]$\label{alg:cfl1}\;
Let $\texttt{Range}(W) = \{0, \cdots, N\}$\;
Let $\mathcal{Y}_w \leftarrow \{y \mid W(x)=w \text{ and } (x, y)\in\mathcal{D}\}$\;
Let $g(y)\leftarrow [\knn(y, \mathcal{Y}_0), \cdots, \knn(y, \mathcal{Y}_N)]$\label{alg:cfl3}\;
Let $T(y_i) \leftarrow \cluster(g(y_1), \cdots, g(y_N))[y_i]$\;
\end{algorithm}\DecMargin{1.5em}

\begin{table}[b!]
\centering
\begin{tabular}{l|cccc}
 & Cold & \elnino{} & \lanina{} & Warm \\
\midrule
EEqt & 2/3 & 0 & 1/3 & 0 \\
WEqt & 0 & 1/4 & 0 & 3/4 \\
EN & $\sim$1/10 & 0 & 1/4 & $\sim$2/3 \\
ES & 3/4 & 0 & 0 & 1/4\\
\end{tabular}
\caption{Each row shows $P(T\mid W=w)$ for a given $w$.}
\label{tab:probtab}
\end{table}

Clustering the conditional expectations gives us the macro-variable class $W(x)$ of each input $x$. By construction~\citep{Chalupka2015a}, we have $p(Y \mid x) = P(Y \mid W(x))$ and by assumption the range of $W$ is small. Instead of checking whether Eq.~\eqref{eq:y_eq} holds for a given pair $y_1, y_2$ over all the $x\in \mathcal{X}$, it is thus enough to check whether $p(y_1 \mid W=w) = p(y_2 \mid W=w)$ for each value $w\in\texttt{Range}(W)$. For each given $w$ we have a subset $\calY_w \subset \calY$ which consists of all the $y$'s whose corresponding $x$'s have causal class $w$. Consequently, Eq.~\eqref{eq:y_eq} does not depend on the exact densities conditional on the micro-state, but only the densities conditional on the macro-level state. Thus, instead of trying to evaluate any given $p(y \mid w)$, Line~\ref{alg:cfl3} computes \emph{the distance of $y$ to the k-th nearest neighbor in $\mathcal{Y}_w$}. This idea is based on a principle that underlies a whole class of nonparametric density estimation algorithms~\citep{Fukunaga1973,Mack1979}: Where the density is high, samples from the distribution are closer to each other than where the density is low. This is illustrated in Fig~\ref{fig:c_vs_cfl}. On the right, we plotted the projection of the data onto the y-space. In this projection, the distance of $y_1$ to its third-nearest neighbor is roughly the same as the distance of $y_2$ to its third-nearest neighbor. Indeed, this is the case for all the $y$'s, because they are generated from a distribution that assigns equal density to all of them.

\begin{figure}
\centering
\includegraphics{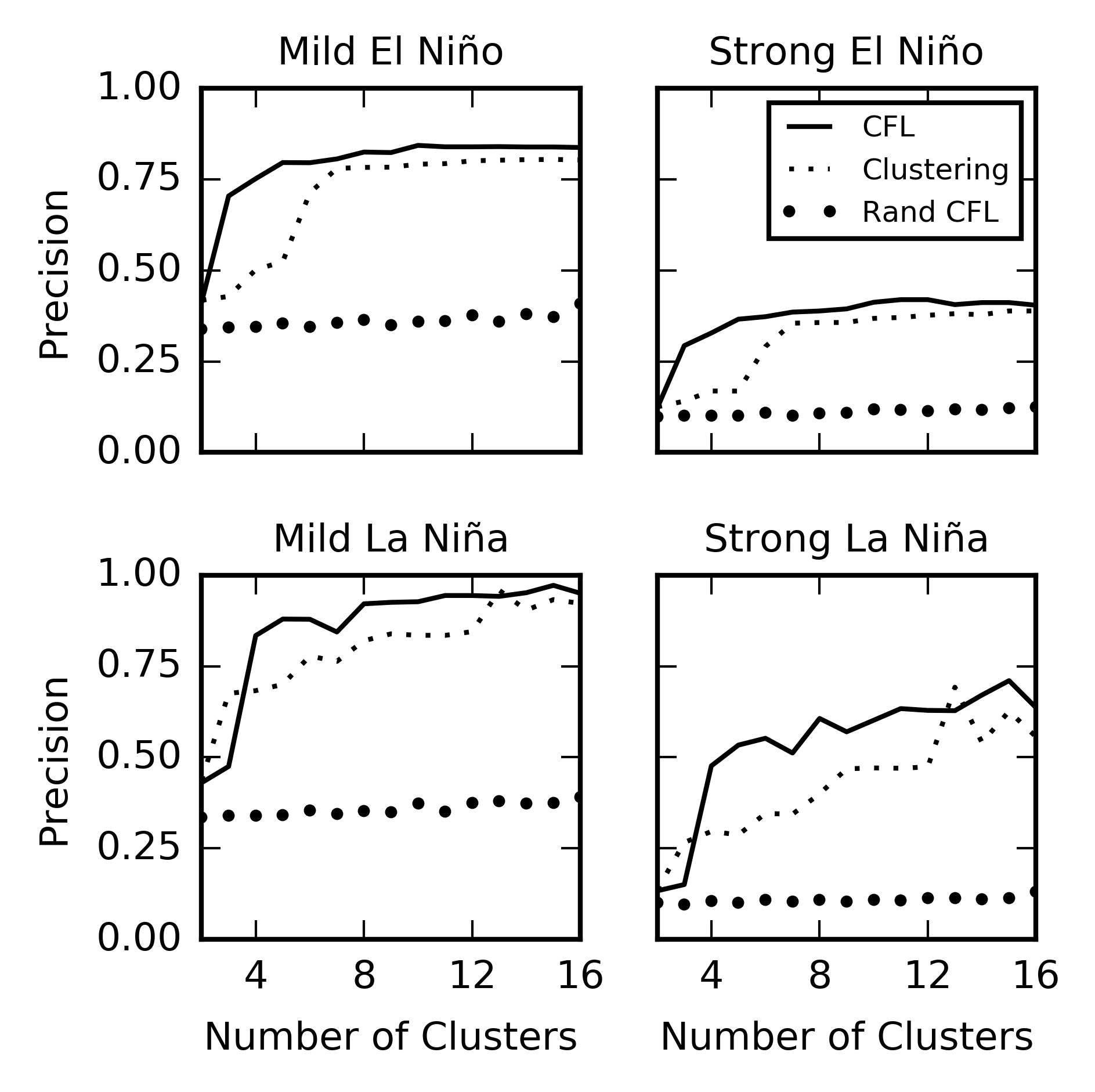}
\caption{Changes in macro-variable precision as we vary the number of states in CFL, clustering, and CFL on reshuffled data (``Rand CFL''). With two states, it is impossible to differentiate \elnino{} and \lanina{} from other weather features, be it dynamic (CFL) or spatio-structural (clustering). Increasing the number of states reveals differences between the algorithms.}
\label{fig:itsnotelnino}
\end{figure}

In \citet{Chalupka2015b} we represented each $y$ by an estimate of $[p(y\mid x_1), \cdots, p(y\mid x_N)]$, where $N$ is the number of datapoints. The new approach represents each $y$ sample by its 'k-nn representation', one scalar value for each $w\in\texttt{Range}(W)$ (Line~\ref{alg:cfl3}). Clustering these representations gives us the causal state $T(y)$ for each $y$. 

Algorithm~\ref{alg:cfl} relies on a successful regression $f$ that minimizes the mean squared error $\mathbb{E}[(f(x)-y)^2]$. In our experiments, we used the Theano~\citep{Theano2012} and Lasagne packages to implement and train a three-hidden-layers, fully-connected neural network~\citep{Bishop1995} in Python. The data was sufficiently simple (compared to e.g.\ image datasets used to evaluate state-of-the-art neural nets in vision) that no regularization technique beyond simple weight decay and early stopping was necessary to minimize the validation error. 

\section{ROBUSTNESS OF THE RESULTS}
In this section, we describe two additional studies we performed to ensure our algorithm behaves as expected, and that the results are robust with respect to changing the experimental parameters.

\subsection{VARYING THE NUMBER OF STATES}
\label{sec:varying_clusters}
Our choice of discovering four-state macro-variables was rather arbitrary. To check how varying the number of states changes the macro-variable precision (Def.~\ref{def:enso_coeff}), we repeated our experimental procedure, varying the number of states K from 2 to 16 (both in the ZW and SST space). Fig.~\ref{fig:itsnotelnino} shows the precisions for each case. As expected, a low number of states (K=2, 3) doesn't allow the algorithm to precisely detect \elnino{} and \lanina{}. With K $>$ 4 however, a slowly growing trend persists at high precision values. \elnino{} and \lanina{} remain important features as K changes.

\begin{figure}
\centering
\includegraphics{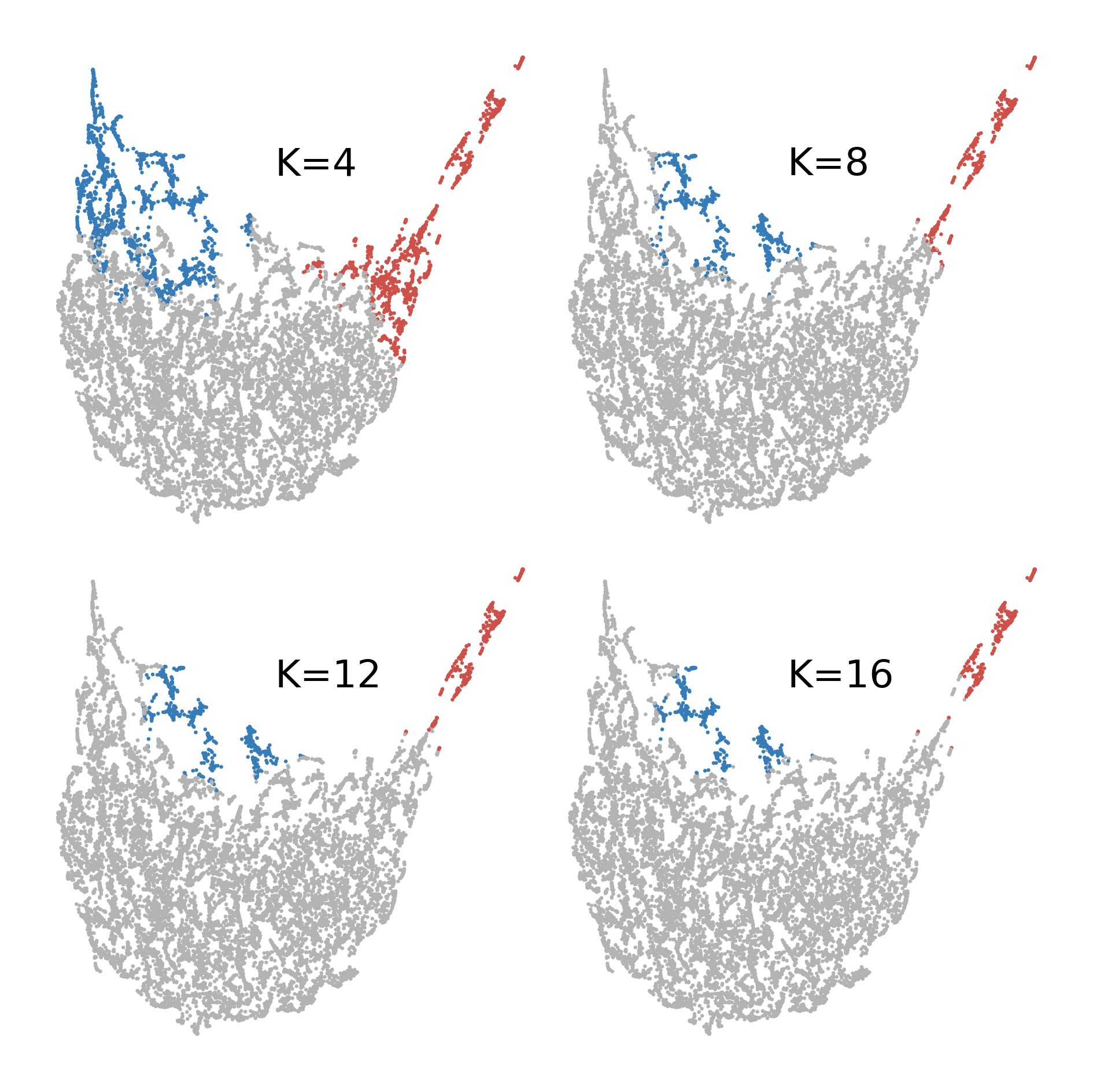}
\caption{t-SNE~\citep{tsne} embedding of the k-nn representation of SST data. The blue dots show, for varying K, the state of T with largest $c_{(-.5)}$ precision (see Def.~\ref{def:enso_coeff}). The red dots show the state with largest $c_{.5}$. Thus, the blue dots are ``the'' \lanina{} cluster for each K, and the red dots ``the'' \elnino{} cluster.}
\label{fig:substates}
\end{figure}

There are several possible behaviors of the algorithm given the slowly growing precision of the macro-variables with growing K: (1) The \elnino{} and \lanina{} states remain roughly constant, (2) CFL sub-divides the \elnino{} and \lanina{} states, (3) CFL finds better \elnino{} and \lanina{} regions, (3) A mix of the above. Fig.~\ref{fig:substates} suggests that (2) is true. As K grows, the clusters that most precisely detect the mild \elnino{} and mild \lanina{} phenomena form a chain of strict subsets. 

\subsection{RESHUFFLED DATA}
\label{sec:sanity}
As a sanity check, we ran Alg.~\ref{alg:cfl} on randomly reshuffled (across the time dimension) ZW and SST data. We asked the algorithm to find K=4, \ldots, 16-state ZW and SST macro-variables. Table~\ref{tab:sanity} shows $P(T \mid W)$, where $W$ and $T$ are the input and output macro-variables discovered in the randomized dataset with $K=4$. Note that $P(T \mid W=W1)$, $P(T \mid W=W2)$, $P(T\mid W=W3)$ and $P(T\mid W=W4)$ are all equal. This is exactly as expected, since by reshuffling the data we removed any probabilistic dependence between the inputs and the outputs. 

Applying Definition~\ref{def:macrovars} to this data indicates that the algorithm implicitly only discovered one true input state, even though we explicitly asked it to look for a four-state macro-variable. The cardinality of the output macro-variable is three or four states, depending on whether $.25$ is close enough to $.27$ to apply Def.~\ref{def:macrovars} to merge the last two columns. We performed the same reshuffled analysis for each $K$ and computed as before the precision for the weak and strong \elnino{} and the weak and strong \lanina{}. Fig.~\ref{fig:itsnotelnino}, large dotted lines, shows that in each case none of the clusters contains a significant proportion of either \elnino{} or \lanina{} patterns.
This experiment offers two insights:

\begin{compactitem}
\item Alg.~\ref{alg:cfl} passes the sanity check. When the inputs and outputs are independent, the input macro-variable is trivial, it has a single state. 
\item When SST patterns are clustered according to their probability of occurrence (e.g.\ as the $W$ variable does in Table~\ref{tab:sanity}), \elnino{} and \lanina{} are not identified as macro-level climate states. We will return to this point in the Discussion.
\end{compactitem}

\begin{table}
\centering
\begin{tabular}{l|cccc}
& T1 & T2 & T3 & T4 \\
\midrule
W1 & .075 & .40 & .25 & .27 \\
W2 & .083 & .39 & .25 & .27 \\
W3 & .084 & .39 & .26 & .27 \\
W4 & .080 & .40 & .24 & .27\\
\end{tabular}
\caption{Conditional probabilities $P(T\mid W)$ when Alg.~\ref{alg:cfl} is applied to randomly (in time) reshuffled ZW and SST data.}
\label{tab:sanity}
\end{table}

\section{WHY NOT NAIVE CLUSTERING?}
\label{sec:clustering}

It is instructive to compare our results with unsupervised clustering. Fig.~\ref{fig:itsnotelnino} shows the precision coefficients for k-means clustering with k=4, \ldots, 16 (small dotted line), alongside our CFL results. Whereas CFL detects both \elnino{} and \lanina{} with high precision using only four states, k-means struggles to achieve a similar result even for larger K.

Barring particularities of the data (which we consider in the Discussion), there is in general no reason for CFL to give the same results as clustering. Consider the example in Fig.~\ref{fig:c_vs_cfl}. Arguably, a reasonable clustering algorithm should find four linearly separable clusters in the joint $\calX, \calY$ space, and two clusters in the $\calX$ and $\calY$ space each. However, the variables are probabilistically independent. In contrast, CFL would only find a one-state input variable, since all values of $X$ imply the same distribution over $Y$. Additionally, since $P(Y \mid X) = P(Y)$ is constant across all the samples, CFL would also only find a one-state output variable. The figure illustrates that Alg.~\ref{alg:cfl} does precisely that (as should the original algorithm in~\citet{Chalupka2015b}).

\section{DISCUSSION}\label{sec:discussion}
The CFL framework we developed in~\citet{Chalupka2015a,Chalupka2015b} aspires to solve an important problem in causal reasoning: how to automatically form macro-level variables from micro-level observations. In this work we have shown, for the first time, that these algorithms can be successfully applied to real-life data. We have recovered well-known, complex climate phenomena (\elnino{}, \lanina{}) as macro-variable states directly from climate data, in an entirely unsupervised manner. In order to do so, we developed a new, practical version of the original CFL algorithm.

We emphasize that our experiments use \emph{observational} climate data, and we have to be cautious about causal conclusions. It is not even clear \emph{a priori} whether the $ZW \rightarrow SST$ causal direction is a reasonable choice: it is known that wind patterns cause changes in SST and it in turn affects the wind by changing the atmospheric pressure. Feedback loops are commonplace in climate dynamics. 

The Causal Coarsening Theorems in~\citet{Chalupka2015a,Chalupka2015b} provide the basis for an efficient learning of causal relationships based on observational macro-variables -- but some experiments are required. In addition, the theorems were only shown to hold for variables that are not subject to feedback. However, we are hopeful that an extension accounting for feedback can be proven. While real climate experiments are generally not feasible, such a theorem would provide the basis to perform large-scale climate experiments with detailed climate models, for example, to check whether \emph{interventionally} shifting from the $W=0$ zonal wind state to $W=1$ in the climate model increases the likelihood of \elnino{} (i.e.\ of SST ending up in state T=1). Connecting the CFL framework with such experiments is an exciting future direction as it would also enable the possibility of using the macro-variables we have found to inform policy that aims to influence climate phenomena.

Our experiments that compare CFL with clustering showed that, as the number of clusters grows, k-means approaches never exceed CFL's precision in detecting \elnino{} and \lanina{}. One explanation for this finding is that while clustering looks for \emph{spatial features} in the data, CFL looks for \emph{relational probabilistic features}. Fig.~\ref{fig:itsnotelnino} suggests that when the number of clusters is small there are strong spatial features in the data that supersede \elnino{} and \lanina{} in their distinctiveness. In contrast, CFL already detects \elnino{} with high precision with only four clusters. This indicates that either (1) There is something unique about $P(\text{\elnino{}} \mid \text{W})$ and $P(\text{\lanina{}} \mid \text{W})$, or (2) There is something unique about $P(\text{\elnino{}})$ and $P(\text{\lanina{}})$. Since we disproved the second hypothesis in Sec.~\ref{sec:sanity}, our results overall indicate that the \elnino{} and \lanina{} phenomena do not only constitute interesting spatial features of the SST map, but are also crucially characterized by the dynamic aspect of the interplay between zonal winds and sea surface temperatures.

Even when working with purely observational data, CFL offers an important causal insight not revealed by clustering methods. It guards against learning variables with ambiguous manipulation effects~\citep{Spirtes2004}. An illustrative example of an ambiguous macro-variable is total cholesterol. Low density lipids (LDL, commonly called ``bad cholesterol'') and high density lipids (HDL, ``good cholesterol'') can be aggregated together to count total cholesterol (TC), but TC has an ambiguous effect on heart disease because effects of LDL and HDL differ. The Causal Coarsening Theorem guarantees that each state of the observational macro-variable is causally unambiguous: no mixing of HDL and LDL can occur. In case of our El Ni\~{n}o setup, this means that two ZW states within the same cell are guaranteed to have the same effect on the SST macro-variable. 

Finally, we note that there still is significant debate among climate scientists about what exactly constitutes \elnino{} and what its causes are. For example, recent research has shown that there may be multiple different types of \elnino{} states \citep{kao2009, johnson2013} that all fall under NOAA's definition. Our results suggest that the current definition described in Section~\ref{sec:elnino} coincides well with states of the probabilistic macro-variable discovered by CFL. In addition, Sec.~\ref{sec:varying_clusters} indicates that finer-grained structure does exist within the \elnino{} and \lanina{} clusters when they are analyzed from the relational-probabilistic standpoint. We leave this line of research as an important future direction.

\subsubsection*{Acknowledgements}
KC’s and PP’s work was supported by the ONR MURI grant N00014-10-1-0933 and Gordon and Betty Moore Foundation.
\subsubsection*{References}
\renewcommand{\section}[2]{}%
\bibliographystyle{plainnat}
\bibliography{bibliography}

\end{document}